%% file: paper.tex
\documentclass[lettersize,journal]{IEEEtran}
\usepackage{amsmath,amsfonts}
\usepackage{algorithmic}
\usepackage{algorithm}
\usepackage{array}
\usepackage[caption=false,font=normalsize,labelfont=rm,textfont=rm]{subfig}
\usepackage{textcomp}
\usepackage{stfloats}
\usepackage{url}
\usepackage{verbatim}
\usepackage{graphicx}
\usepackage{cite}
\usepackage{multirow}
\usepackage{amsfonts,amssymb}
\usepackage{enumerate}
\usepackage{color}
\usepackage[table,xcdraw,dvipsnames,svgnames,x11names]{xcolor}
\usepackage[normalem]{ulem}
\useunder{\uline}{\ul}{}
\usepackage{pifont}
\usepackage{booktabs}
\usepackage{hyperref}
\usepackage{adjustbox}
\DeclareSymbolFont{mathbf}{OT1}{cmr}{bx}{n}
\DeclareMathSymbol{X}{\mathalpha}{mathbf}{`X}
\DeclareMathSymbol{Z}{\mathalpha}{mathbf}{`Z}
\DeclareMathSymbol{Y}{\mathalpha}{mathbf}{`Y}
\DeclareMathSymbol{W}{\mathalpha}{mathbf}{`W}
\DeclareMathSymbol{E}{\mathalpha}{mathbf}{`E}
\begin{document}

\title{JL1-CC\&QA: Extending the JL1-CD Benchmark with Change Captioning and Question Answering}

\author{Ziyuan Liu, Ruifei Zhu, Ouqiao Ma, and Yuantao Gu
\thanks{Ziyuan Liu and Yuantao Gu are with the Department of Electronic Engineering, Beijing National Research Center for Information Science and Technology, Tsinghua University, Beijing 100084, China (e-mail: liuziyua22@mails.tsinghua.edu.cn; gyt@tsinghua.edu.cn).
Ruifei Zhu is with Chang Guang Satellite Technology Co., Ltd. (CGSTL)
Changchun 130102, China (e-mail: zhuruifei@jl1.cn).
Ouqiao Ma is with the College of Communications
Engineering, Army Engineering University of PLA, Nanjing 210007,
China (e-mail: moq@aeu.edu.cn).
(\textit{Corresponding author: Yuantao Gu.})}
}

\maketitle
\begin{abstract}
Remote sensing change detection (CD) traditionally focuses on pixel-level binary segmentation, which identifies \textit{where} changes occur but neither \textit{what} nor \textit{why}.
To bridge this semantic gap, we introduce {JL1-CC\&QA}, a multi-task benchmark that extends the JL1-CD dataset with two complementary annotation layers: change captioning (CC) and change question answering (QA). Built upon 5{,}000 bi-temporal image pairs acquired by the Jilin-1 satellite at 0.5--0.75\,m ground sample distance, the benchmark comprises:
(i)~\textbf{JL1-CC}, providing 17{,}021 quality-verified captions that describe diverse land-cover transformations; and
(ii)~\textbf{JL1-QA}, offering 20{,}060 question--answer pairs across eight question types, enabling fine-grained, interactive interrogation of surface changes.
All annotations are produced via a three-stage pipeline consisting of multi-modal large language model (LLM) generation, vision-grounded LLM judging, and human expert verification.
We hope that JL1-CC\&QA, as a benchmark unifying binary change masks, change captions, and change-oriented QA over the same image set, will serve as a valuable resource for the community to advance multi-task change understanding in remote sensing.
The dataset is available at \url{https://github.com/circleLZY/JL1-CD}.

\end{abstract}


\section{Introduction}
\label{sec:introduction}

\IEEEPARstart{R}{emote} sensing change detection (CD) aims to identify surface changes from multi-temporal imagery acquired over the same geographic area, serving as an important technology for urban planning, environmental monitoring, disaster response, and resource management~\cite{lv2022land, bai2023deep}. Over the past decade, the rapid growth of both Earth observation data and deep learning methods has propelled CD into one of the most active research frontiers in remote sensing.

The predominant formulation of CD is binary change detection (BCD), which classifies each pixel as either \textit{changed} or \textit{unchanged}. A large number of BCD benchmarks have been established, spanning diverse geographic contexts, spatial resolutions, and change categories (see Table~\ref{tab:datasets} for a comprehensive summary)~\cite{benedek2009change, bourdis2011constrained, daudt2018urban, lebedev2018change, ji2019fully, wang2019getnet, lopezfandino2019gpu, chen2020stanet, zhang2020ifn, peng2021semicdnet, shen2021s2looking, shao2021sunet, shi2022sysucd, liu2022clcd, li2022mscdunet, holail2023afdenet, liu2025jl1cd}. While the vast majority of these benchmarks adopt the {bi-temporal} setting and have driven extensive algorithmic development~\cite{daudt2018fully, chen2020stanet, zhang2020ifn, fang2022snunet, chen2022bit, bandara2022changeformer, chen2024changemamba}, the community has also explored {single-temporal} CD that exploits pretrained semantic representations to infer changes from unpaired imagery~\cite{zheng2021changestar, chen2023i3pe, zhou2024stmnet}, as well as {multi-temporal} monitoring that tracks continuous change trajectories from dense satellite time series~\cite{toker2022dynamicearthnet, vanetten2021spacenet7, shi2023wusu, zhao2024coud}. Despite these differences, all of the above approaches produce pixel-level masks that encode \textit{where} change occurred but remain silent on \textit{what} changed or \textit{why}.

\input{table/benchmarks}

In parallel, the range of input modalities has expanded considerably. While optical RGB imagery remains the dominant data source for CD research~\cite{chen2020stanet, ji2019fully, shen2021s2looking, shi2022sysucd, liu2025jl1cd}, multispectral and hyperspectral sensors offer richer spectral discrimination for fine-grained change analysis~\cite{daudt2018urban, wang2019getnet, lopezfandino2019gpu, hu2021acda, wang2022sstformer, xie2024dfgan, dong2025ct2net}, and cross-modal fusion of optical and SAR data enables all-weather, day-and-night monitoring~\cite{he2023cauflood, li2021dtcdn, li2022mscdunet, shao2021sunet, alatalo2023improved, liu2025m2cd, shen2025sdmamba}. Although these modalities provide richer visual information, the output of BCD remains a pixel-level binary mask devoid of semantic content.

Semantic change detection (SCD) and building damage assessment (BDA) partially bridge this gap by assigning categorical labels to changed pixels. SCD introduces per-pixel land-cover annotations at both time phases, enabling identification of change {types} (e.g., farmland$\to$buildings)~\cite{daudt2019multitask, yang2022second, tian2022hiucd, ding2022bisrnet, yuan2022landsat, pang2023bandon, shi2023wusu, tang2024clearscd, liu2024cropscd, tan2025triples, fang2025prohrscd}. BDA further introduces ordinal damage scales (e.g., minor/major/destroyed) for disaster response~\cite{fujita2017damage, gupta2019xbd, zheng2021changeos, rahnemoonfar2023rescuenet, sun2024qqb, chen2025bright, wang2025cdfnet}. Nevertheless, the semantic content in these datasets is still encoded as {discrete numerical class indices} drawn from a closed taxonomy. This gap calls for a shift toward natural-language-grounded change understanding.

The advent of vision--language models (VLMs) has begun to close this gap. In the natural image domain, change captioning benchmarks first demonstrated the feasibility of describing visual differences in natural language~\cite{jhamtani2018spotthediff, park2019clevrchange}. The remote sensing community quickly adopted this paradigm, producing a growing family of change captioning (CC) datasets that range from low-resolution multitemporal pairs to very-high-resolution urban and disaster scenes~\cite{hoxha2022dubaiccd, liu2022levircc, liu2024levirMCI, li2025secondcc, chen2024ccexpert, li2025rscc, turker2025mosaicsen2cc}. In parallel, change detection visual question answering (CDVQA) has emerged as a complementary task, enabling users to query specific aspects of surface changes through natural-language questions~\cite{yuan2022cdvqa, li2024qag360k}, while instruction-tuning datasets have further extended the interaction to multi-turn conversational analysis~\cite{wu2025changechat, irvin2025teochatlas, wang2025disasterm3}. Despite this rapid progress, rare benchmarks jointly provide binary change masks, change captions, and change-oriented question--answer pairs on the same set of image pairs, which limits multi-task learning and cross-task evaluation.

\begin{figure*}[!t]
    \centering
    \includegraphics[width=0.95\textwidth]{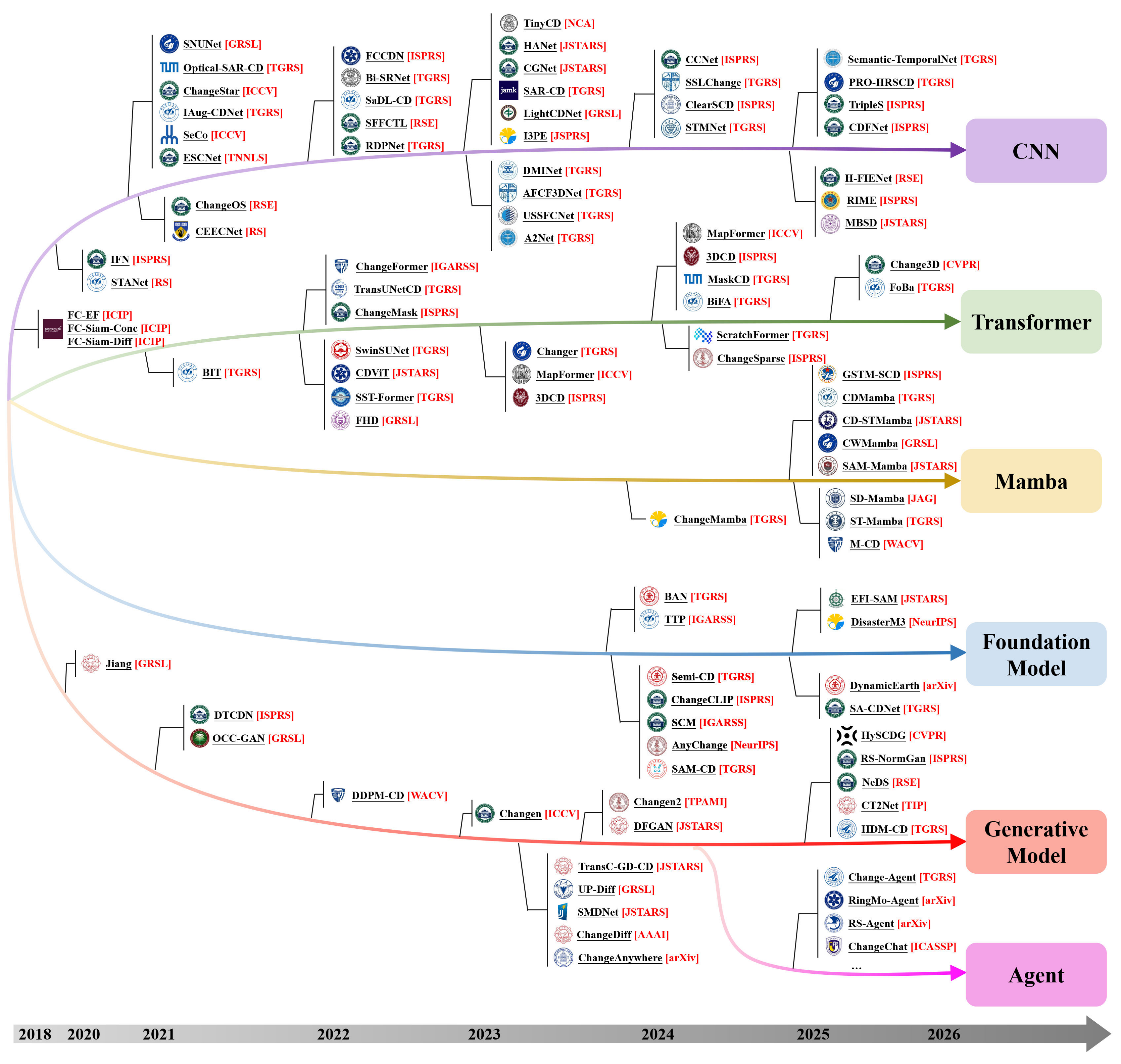}
    \caption{Timeline of the development of mainstream deep learning-based CD methods.}
    \label{fig:timeline}
\end{figure*}

The evolution of datasets has been accompanied by corresponding advances in model architectures (see Fig.~\ref{fig:timeline}). Methods built on {CNN} backbones with change decoders~\cite{daudt2018fully, chen2020stanet, zhang2020ifn, fang2022snunet, chen2022fccdn, chen2022rdpnet, chen2022sadlcd, chen2021iaugcdnet, li2023a2net, codegoni2023tinycd, xing2023lightcdnet, lei2023ussfcnet, zhang2023escnet, feng2023dminet, ye2023afcf3dnet, han2023hanet, han2023cgnet, cao2023ffctl, cheng2024ccnet, zhao2024sslchange, zheng2024changesparse, liu2025mbsd, sun2025riem, sun2025semantictemporalnet, mbsd, UnlearningCD} established the dominant Siamese paradigm, while {Transformer}~\cite{chen2022bit, bandara2022changeformer, zhang2022swinsunet, li2022transunetcd, shi2022cdvit, zheng2022changemask, noman2024scratchformer, fang2023changer, bernhard2023mapformer, wang2022sstformer, yu2024maskcd, zhang2024bifa, zhang2025foba, zhu2025change3d} introduced global cross-temporal attention, and {Mamba}-based methods~\cite{chen2024changemamba, zhang2025cdmamba, liu2025cdstmamba, zhao2025stmamba, liu2025cwmamba, liu2025gstmscd, li2025sammamba, shen2025sdmamba, paranjape2025mcd} achieved comparable perception at linear computational cost.
{Foundation model} adaptations have enabled zero-shot and few-shot CD by transferring pretrained CLIP and SAM representations~\cite{dong2024changeclip, zheng2024anychange, ding2024samcd, li2024ban, tan2024scm, li2024semicdvl, huang2025efisam, qin2025sam2cd, li2025dynamicearth}.
{Generative} approaches based on GANs~\cite{jiang2020semisupervised, jian2022occgan, li2021dtcdn, zheng2023changen} and diffusion models~\cite{bandara2022ddpmcd, zheng2024changen2, tang2024changeanywhere, jia2024smdnet, han2025hdmcd, zang2025changediff, benidir2025hyscdg} have addressed data scarcity through synthetic bi-temporal image generation.
{Agent}-based systems have integrated LLMs as reasoning engines for interactive change analysis~\cite{liu2024changeagent, xu2024rsagent, feng2025earthagent, hu2025ringmoagent}. A critical observation is that the {text modality} has been a key enabler at each stage: CLIP-based models use text prompts to guide change semantics, captioning models translate visual change into language, VQA models support interactive querying, and agents conduct multi-step reasoning in natural language. This trajectory underscores the need for language-grounded benchmarks that can support the training and evaluation of these increasingly capable architectures.

Motivated by the above analysis, we present \textbf{JL1-CC\&QA}, extending the JL1-CD benchmark~\cite{liu2025jl1cd}, which consists of 5{,}000 bi-temporal image pairs from the Jilin-1 satellite with a resolution of 0.5--0.75\,m, with two new annotation layers: (i)~\textbf{JL1-CC}, providing 17{,}021 quality-verified change captions spanning both anthropogenic and natural changes; and (ii)~\textbf{JL1-QA}, offering 20{,}060 question--answer pairs across eight question types (existence, description, location, magnitude, temporal comparison, causation, relative comparison, and visual detail). All annotations are produced via a three-stage pipeline: multi-modal LLM generation, vision-grounded LLM judging, and human expert verification. Our main contributions are summarized as follows:

\begin{enumerate}
    \item We construct a change understanding benchmark that provides binary change masks, natural-language change descriptions, and diverse QA pairs over 5{,}000 bi-temporal satellite image pairs, facilitating joint training and evaluation across CD, CC, and QA tasks.
    \item We design a scalable, reproducible annotation pipeline that couples multi-modal LLM generation with vision-grounded LLM judging and human verification.
    \item We provide comprehensive dataset statistics, and release all data and code to support community research toward unified remote sensing change understanding.
\end{enumerate}


\section{Dataset Construction}
\label{sec:dataset}

This section describes the construction of JL1-CC\&QA. We first introduce the source dataset JL1-CD (Section~\ref{subsec:jl1cd}), then detail the change captioning pipeline (Section~\ref{subsec:cc_pipeline}) and the question answering pipeline (Section~\ref{subsec:qa_pipeline}).

\subsection{JL1-CD}
\label{subsec:jl1cd}

JL1-CC\&QA is built upon JL1-CD~\cite{liu2025jl1cd}, which comprises 5{,}000 bi-temporal image pairs captured by the Jilin-1 high-resolution optical satellite. The imagery was acquired across multiple provinces in China, including Shandong, Ningxia, Anhui, Hebei, and Hunan, between early 2022 and late 2023, and was carefully curated to exclude blur, cloud occlusion, and extreme illumination conditions. Each image pair consists of a pre-event image, a post-event image, and a pixel-level binary change mask annotated by professional interpreters. Key specifications are summarized in Table~\ref{tab:jl1cd_specs}.

\begin{table}[t]
\centering
\caption{Specifications of the JL1-CD Source Dataset}
\label{tab:jl1cd_specs}
\footnotesize
\renewcommand{\arraystretch}{1.2}
\begin{tabular}{ll}
\toprule
\textbf{Attribute} & \textbf{Value} \\
\midrule
Satellite        & Jilin-1 (JL1) Constellation \\
Spatial resolution & 0.5--0.75\,m GSD \\
Image size       & $512 \times 512$ pixels \\
Spectral bands   & RGB (3 channels) \\
Total image pairs & 5{,}000 \\
Training / test split & 4{,}000 / 1{,}000 \\
Annotation       & Pixel-level binary mask \\
Temporal coverage & 2022--2023 \\
Geographic coverage & Multiple provinces in China \\
\bottomrule
\end{tabular}
\end{table}

\begin{figure}[!t]
    \centering
    \includegraphics[width=0.80\columnwidth]{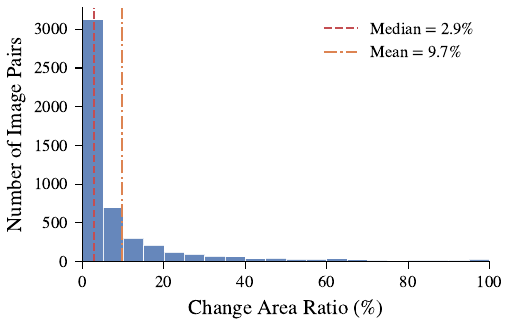}
    \caption{Distribution of change area ratio (CAR) across 5{,}000 image pairs in JL1-CD.}
    \label{fig:car_distribution}
\end{figure}

Two properties of JL1-CD make it suitable for change captioning and question answering. First, the dataset is {all-inclusive} in terms of change types: it encompasses both anthropogenic changes (buildings, roads, hardened surfaces, photovoltaic panels, etc.) and natural changes (woodlands, grasslands, croplands, water bodies, etc.). This diversity ensures that the derived CC and QA annotations cover a broad spectrum of real-world surface dynamics. Second, the change area ratio (CAR)---defined as the proportion of changed pixels in each image pair---spans the full range from near-zero to 100\%, with a mean of 9.7\% and a median of 2.9\% (Fig.~\ref{fig:car_distribution}). This long-tailed distribution mirrors the natural imbalance of real-world change scenarios and poses a meaningful challenge for language-grounded change understanding at varying scales.

\subsection{JL1-CC}
\label{subsec:cc_pipeline}

\subsubsection{Task Definition}
Given a bi-temporal image pair $(I_A, I_B)$ and its corresponding binary change mask $M$, the change captioning task requires generating a set of natural-language sentences $\{c_1, c_2, \ldots, c_k\}$ that accurately describe the semantic content of the observed surface changes, including the type, location, and extent of change.

\subsubsection{Annotation Pipeline}
\begin{figure*}[!t]
    \centering
    \includegraphics[width=0.95\textwidth]{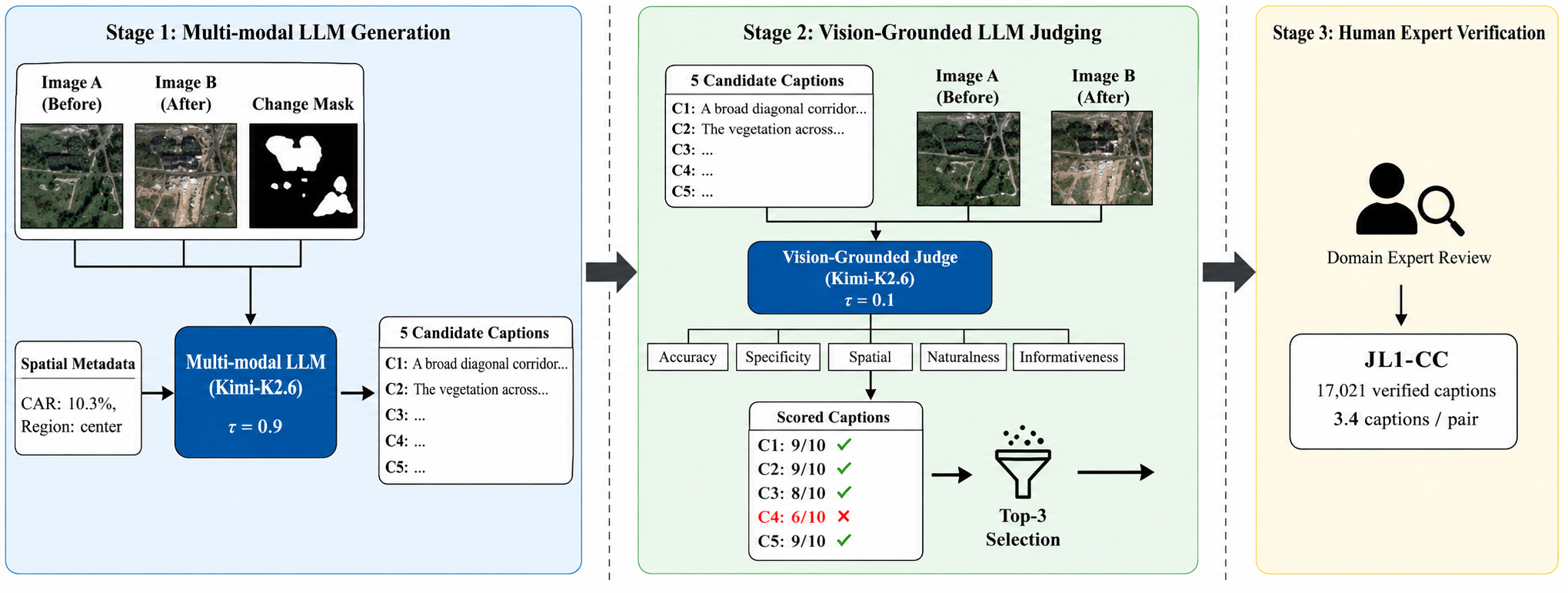}
    \caption{Overview of the JL1-CC annotation pipeline. Stage~1: a multi-modal LLM generates five candidate captions from the image pair and spatial metadata. Stage~2: a vision-grounded LLM judge scores each caption and retains the top three. Stage~3: human experts verify factual accuracy.}
    \label{fig:cc_pipeline}
\end{figure*}

As illustrated in Fig.~\ref{fig:cc_pipeline}, the JL1-CC annotation pipeline consists of three stages.

\textbf{Stage~1: Multi-modal LLM Generation.}
For each image pair, we prompt a multi-modal LLM (Kimi-K2.6) with three visual inputs: the pre-event image $I_A$, the post-event image $I_B$, and the binary change mask $M$, together with spatial metadata including the CAR and a textual description of the primary change region (e.g., ``upper-left area of the image''). The model is instructed to generate five diverse captions, each describing the same change from a different perspective: change type, spatial location, visual appearance, scale, or implication.

\textbf{Stage~2: Vision-Grounded LLM Judging.}
A second LLM call evaluates each caption by examining the original image pair alongside the generated text. Each caption is scored on a 1--10 integer scale across five criteria: (1)~accuracy: whether the description matches the visible changes, with heavy penalties for hallucination; (2)~specificity: whether concrete land-cover terms are used; (3)~spatial correctness: whether location references are accurate; (4)~naturalness: whether the English is fluent; and (5)~informativeness: whether the caption conveys meaningful detail. The top-3 scoring captions are retained, and ties at the cutoff are preserved.

\textbf{Stage~3: Human Expert Verification.}
A subset of the generated captions is reviewed by domain experts to verify factual accuracy and identify systematic errors, ensuring that the automated pipeline produces reliable annotations at scale.

\subsubsection{Statistics}

\begin{figure}[!t]
    \centering
    \includegraphics[width=0.80\columnwidth]{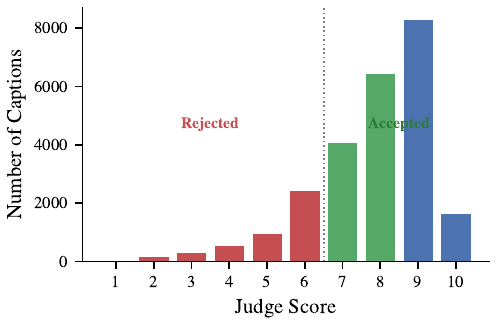}
    \caption{Judge score distribution for JL1-CC. Captions scoring below 7 (red) are rejected; those scoring 7--8 (green) and 9--10 (blue) are retained.}
    \label{fig:cc_score}
\end{figure}

\begin{figure}[!t]
    \centering
    \includegraphics[width=0.90\columnwidth]{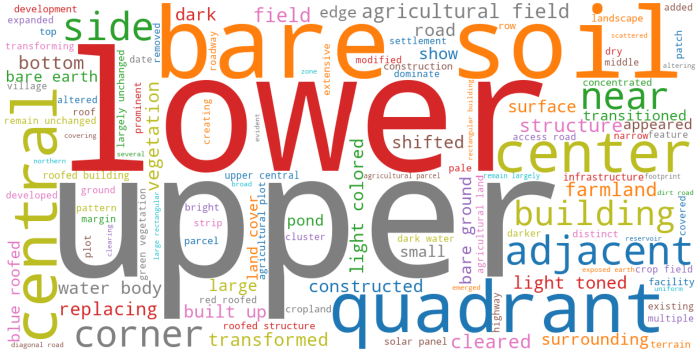}
    \caption{Word cloud of the 17{,}021 selected captions in JL1-CC.}
    \label{fig:cc_wordcloud}
\end{figure}

Table~\ref{tab:cc_stats} summarizes the JL1-CC dataset statistics. The pipeline generates 25{,}000 candidate captions (5 per pair) and retains 17{,}021 after quality filtering, yielding a pass rate of 68.1\%. The judge score distribution (Fig.~\ref{fig:cc_score}) shows clear discrimination: 39.8\% of captions score 9--10, 42.3\% score 7--8, and 17.9\% score below 7 and are rejected. The selected captions have a mean length of 26.2 words, with a vocabulary of 7{,}458 unique tokens. Fig.~\ref{fig:cc_wordcloud} presents a word cloud of the selected captions, where the most prominent terms---\textit{upper}, \textit{lower}, \textit{bare soil}, \textit{agricultural}, \textit{building}, \textit{road}, \textit{water}---reflect both the spatial referencing style and the diversity of land-cover types in JL1-CD.

\begin{table}[t]
\centering
\caption{JL1-CC Dataset Statistics}
\label{tab:cc_stats}
\footnotesize
\renewcommand{\arraystretch}{1.2}
\begin{tabular}{lrrr}
\toprule
\textbf{Statistic} & \textbf{Train} & \textbf{Test} & \textbf{Total} \\
\midrule
Image pairs               & 4{,}000  & 1{,}000  & 5{,}000  \\
Generated captions        & 20{,}000 & 5{,}000  & 25{,}000 \\
Selected captions         & 13{,}616 & 3{,}405  & 17{,}021 \\
Avg.\ selected / pair     & 3.40     & 3.40     & 3.40     \\
Avg.\ caption length (words) & 26.2  & 26.4     & 26.2     \\
Vocabulary size           & 6{,}837  & 4{,}064  & 7{,}458  \\
Judge score (mean / median) & 7.8 / 8.0 & 7.8 / 8.0 & --- \\
\bottomrule
\end{tabular}
\end{table}


\subsection{JL1-QA}
\label{subsec:qa_pipeline}

\subsubsection{Task Definition}
Given a bi-temporal image pair $(I_A, I_B)$ and a natural-language question $Q$, the change question answering task requires generating a textual answer $A$ that accurately responds to the question based on the visible surface changes.

\subsubsection{Question Taxonomy}
We define eight question types adapted for open-ended answer generation, as summarized in Table~\ref{tab:qa_types}. Each type targets a distinct aspect of change understanding, ranging from binary existence judgments to causal reasoning, ensuring comprehensive coverage of user information needs.

\begin{table}[t]
\centering
\caption{Question Types in JL1-QA with Examples}
\label{tab:qa_types}
\footnotesize
\renewcommand{\arraystretch}{1.2}
\begin{tabular}{lp{5.5cm}}
\toprule
\textbf{Type} & \textbf{Example Question} \\
\midrule
YES/NO       & Has the vegetation in the lower half been removed? \\
WHAT         & What happened to the farmland in the center? \\
WHERE        & Where did the most significant change occur? \\
HOW MUCH     & Is the change large-scale or localized? \\
BEFORE/AFTER & What was present in the upper-left before the change? \\
CAUSE        & What type of development likely caused the changes? \\
DETAIL       & What do the new structures appear to be? \\
COMPARE      & Which area shows the most dramatic change? \\
\bottomrule
\end{tabular}
\end{table}

\subsubsection{Annotation Pipeline}
\begin{figure*}[!t]
    \centering
    \includegraphics[width=0.95\textwidth]{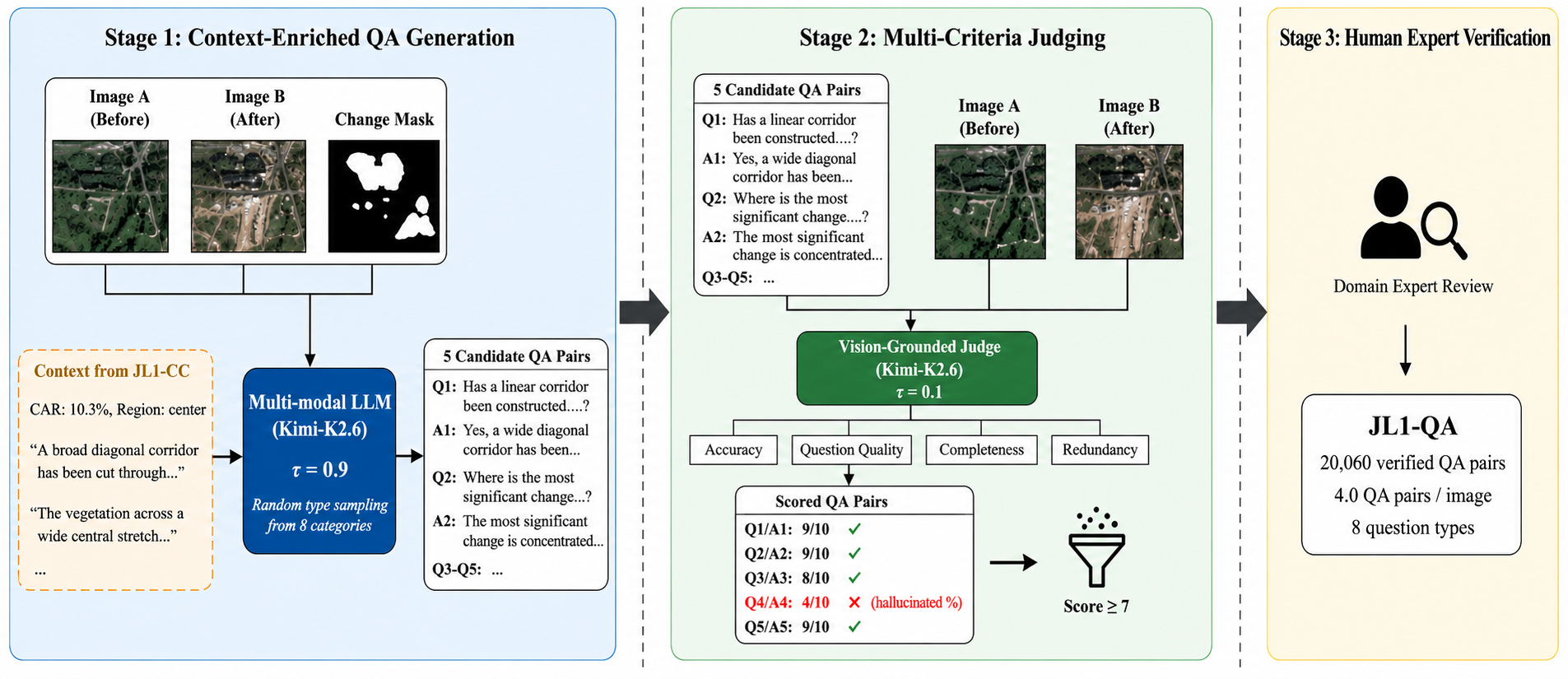}
    \caption{Overview of the JL1-QA annotation pipeline. The generation stage receives the image pair, change mask, and JL1-CC metadata (captions, CAR, change region) as context. The judge evaluates each QA pair for accuracy, quality, completeness, and redundancy.}
    \label{fig:qa_pipeline}
\end{figure*}

As illustrated in Fig.~\ref{fig:qa_pipeline}, the JL1-QA annotation pipeline shares the same three-stage architecture as JL1-CC, with two key enhancements in the generation stage.

\textbf{Stage~1: Context-Enriched QA Generation.}
Beyond the three visual inputs ($I_A$, $I_B$, $M$), the LLM additionally receives contextual metadata from JL1-CC: the change area ratio, the spatial change region description, and up to three selected change captions. This context enrichment grounds the QA generation in verified change descriptions, yielding more accurate and diverse question--answer pairs. The model generates five QA pairs per image, each covering a different question type randomly sampled from the taxonomy. Crucially, the prompt explicitly prohibits copying exact numerical metadata (e.g., change percentages) into answers, as such precision cannot be derived from visual inspection alone.

\textbf{Stage~2: Multi-Criteria Judging.}
Each QA pair is scored on a 1--10 integer scale across four criteria: (1)~answer accuracy: whether the answer is factually correct given the images; (2)~question quality: whether the question is clear, natural, and non-trivial; (3)~answer completeness: whether the answer adequately addresses the question; and (4)~redundancy: whether the QA pair is substantially different from the others for the same image. QA pairs scoring below 7 are discarded. The most common rejection reasons are hallucinated precise percentages (score 1--4), redundancy with other QA pairs (score 5--6), and vague answers (score 5--6).

\textbf{Stage~3: Human Expert Verification.}
As with JL1-CC, domain experts review a subset of selected QA pairs to verify factual accuracy and check for systematic errors in question formulation or answer content.

\subsubsection{Statistics}

\begin{figure}[!t]
    \centering
    \includegraphics[width=0.80\columnwidth]{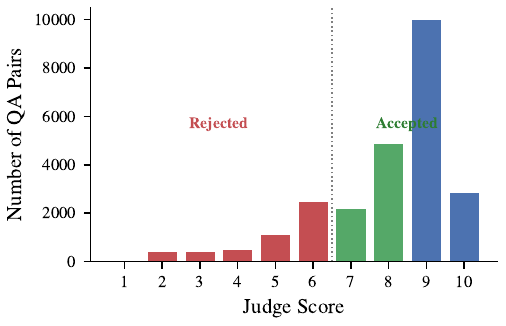}
    \caption{Judge score distribution for JL1-QA. QA pairs scoring below 7 are rejected.}
    \label{fig:qa_score}
\end{figure}

\begin{figure}[!t]
    \centering
    \includegraphics[width=0.80\columnwidth]{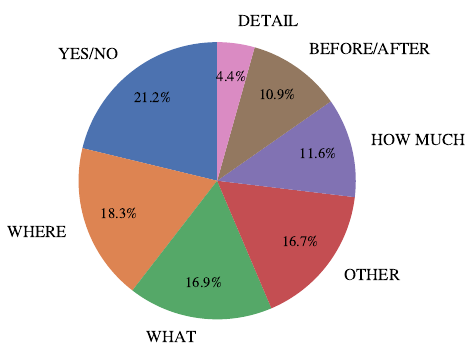}
    \caption{Distribution of question types in the 20{,}060 selected QA pairs of JL1-QA.}
    \label{fig:qa_types}
\end{figure}

Table~\ref{tab:qa_stats} summarizes the JL1-QA dataset statistics. From 24{,}995 generated QA pairs, 20{,}060 pass the quality threshold (score $\geq$ 7), yielding a pass rate of 80.3\%. The higher pass rate compared to JL1-CC (68.1\%) is attributed to the context enrichment from change captions, which reduces hallucination in the generation stage. The judge score distribution (Fig.~\ref{fig:qa_score}) confirms effective quality discrimination: 51.4\% of QA pairs score 9--10, 28.4\% score 7--8, and 20.2\% are rejected. The question type distribution (Fig.~\ref{fig:qa_types}) shows broad coverage across all eight categories, with YES/NO (21.2\%), WHERE (18.3\%), and WHAT (16.9\%) being the most frequent. Questions average 11.1 words and answers 19.3 words in length.

\begin{table}[t]
\centering
\caption{JL1-QA Dataset Statistics}
\label{tab:qa_stats}
\footnotesize
\renewcommand{\arraystretch}{1.2}
\begin{tabular}{lrrr}
\toprule
\textbf{Statistic} & \textbf{Train} & \textbf{Test} & \textbf{Total} \\
\midrule
Image pairs                  & 3{,}999  & 1{,}000  & 4{,}999  \\
Generated QA pairs           & 19{,}995 & 5{,}000  & 24{,}995 \\
Selected QA pairs            & 16{,}055 & 4{,}005  & 20{,}060 \\
Avg.\ selected / pair        & 4.01     & 4.01     & 4.01     \\
Avg.\ question length (words)& 11.1     & 11.1     & 11.1     \\
Avg.\ answer length (words)  & 19.3     & 19.3     & 19.3     \\
Judge score (mean / median)  & 7.9 / 9.0 & 7.9 / 9.0 & ---    \\
\bottomrule
\end{tabular}
\end{table}


\section{Conclusion}
\label{sec:conclusion}

In this paper, we presented JL1-CC\&QA, a multi-task benchmark that extends the JL1-CD binary change detection dataset with two complementary annotation layers: change captioning (JL1-CC) and change question answering (JL1-QA). Both layers are produced through a three-stage pipeline---multi-modal LLM generation, vision-grounded LLM judging, and human expert verification---balancing scalability with factual reliability. We hope this benchmark serves as a valuable resource for the community to advance multi-task change understanding in remote sensing.

\bibliographystyle{IEEEtran}
\bibliography{refs_2}

\end{document}

%% file: table/benchmarks.tex

\begin{table*}[!t]
\centering
\caption{Comprehensive Summary of Representative Remote Sensing Change Detection Datasets}
\label{tab:datasets}
\begin{adjustbox}{max width=\textwidth}
\footnotesize
\setlength{\tabcolsep}{3pt}
\renewcommand{\arraystretch}{1.26}
\begin{tabular}{lllllllllp{4.5cm}}
\toprule
\textbf{Dataset} & \textbf{Year} & \textbf{Task} & \textbf{Modal.} & \textbf{Source} & \textbf{GSD} & \textbf{Size (px)} & \textbf{Scale} & \textbf{Temp.} & \textbf{Change Elements} \\
\midrule
\href{http://web.eee.sztaki.hu/remotesensing/airchange_benchmark.html}{SZTAKI}~\cite{benedek2009change}
  & 2009 & BCD & Opt. & Aerial & 1.5\,m & $952\!\times\!640$ & 13 & Bi & Buildings, forest, foundations \\
\href{https://computervisiononline.com/dataset/1105138664}{AICD}~\cite{bourdis2011constrained}
  & 2011 & BCD & Opt. & Synthetic & --- & $800\!\times\!600$ & 1000 & Bi & Synthetic (buildings, vehicles) \\
\href{https://github.com/gistairc/ABCDdataset}{ABCD}~\cite{fujita2017damage}
  & 2017 & BDA & Opt. & Aerial & 0.4\,m & $128\!\times\!128$ & 8506 & Bi & Building damage \\
\href{https://github.com/rulixiang/MtS-WH-Dataset/blob/master/README_CHN.md}{MtS-WH}~\cite{wu2017kernel}
  & 2017 & BCD & Opt. & IKONOS & 1\,m & $7200\!\times\!6000$ & 1 & Bi & Parking, water, bldg., farm, industrial \\
\href{https://ieee-dataport.org/open-access/oscd-onera-satellite-change-detection\#files}{OSCD}~\cite{daudt2018urban}
  & 2018 & BCD & MS & Sentinel-2 & 10\,m & $600\!\times\!600$ & 24 & Bi & Buildings, roads \\
\href{https://drive.google.com/file/d/1GX656JqqOyBi_Ef0w65kDGVto-nHrNs9/edit}{CDD}~\cite{lebedev2018change}
  & 2018 & BCD & Opt. & GE & VHR & $256\!\times\!256$ & 16000 & Bi & Seasonal, vehicles, bldg., trees \\
\href{http://study.rsgis.whu.edu.cn/pages/download/}{WHU-CD}~\cite{ji2019fully}
  & 2018 & BCD & Opt. & Aerial & 0.2\,m & $15354\!\times\!32507$ & 1 & Bi & Buildings \\
\href{https://ieee-dataport.org/open-access/hrscd-high-resolution-semantic-change-detection-dataset\#files}{HRSCD}~\cite{daudt2019multitask}
  & 2019 & SCD & Opt. & Aerial (IGN) & 0.5\,m & $10000\!\times\!10000$ & 291 & Bi & Artificial, agriculture, forest, wetland, water \\
\href{https://drive.google.com/file/d/1cWy6KqE0rymSk5-ytqr7wM1yLMKLukfP/view}{River}~\cite{wang2019getnet}
  & 2019 & BCD & HS & Airborne & --- & $463\!\times\!241$ & 1 & Bi & Water \\
\href{https://gitlab.citius.gal/hiperespectral/ChangeDetectionDataset}{Hyper.\ CDD}~\cite{lopezfandino2019gpu}
  & 2019 & BCD & HS & Airborne & --- & Varies & 3 & Bi & Crop transition \\
\href{https://github.com/DIUx-xView/xView2_baseline?tab=readme-ov-file}{xBD}~\cite{gupta2019xbd}
  & 2019 & BDA & Opt. & Satellite & VHR & $1024\!\times\!1024$ & 22068 & Bi & Building damage \\
\href{https://justchenhao.github.io/LEVIR/}{LEVIR-CD}~\cite{chen2020stanet}
  & 2020 & BCD & Opt. & GE & 0.5\,m & $1024\!\times\!1024$ & 637 & Bi & Buildings \\
\href{https://github.com/S2Looking/Dataset/tree/main/LEVIR-CD\%2B/LEVIR-CD\%2B}{LEVIR-CD+}~\cite{chen2020stanet}
  & 2020 & BCD & Opt. & GE & 0.5\,m & $1024\!\times\!1024$ & 985 & Bi & Buildings \\
\href{https://github.com/GeoZcx/A-deeply-supervised-image-fusion-network-for-change-detection-in-remote-sensing-images/tree/master/dataset}{DSIFN}~\cite{zhang2020ifn}
  & 2020 & BCD & Opt. & GE & VHR & $512\!\times\!512$ & 3940 & Bi & Roads, bldg., farmland, water \\
\href{https://github.com/daifeng2016/Change-Detection-Dataset-for-High-Resolution-Satellite-Imagery}{CD\_Data\_GZ}~\cite{peng2021semicdnet}
  & 2020 & BCD & Opt. & GE & VHR & $256\!\times\!256$ & 4000+ & Bi & Water, road, farmland, bare soil, forest, bldg., ships \\
Optical-SAR-CD~\cite{saha2022selfsupervised}
  & 2021 & BCD & Cross & Opt.+SAR & 10\,m & $600\!\times\!600$ & 3 & Bi & Mixed land cover \\
\href{https://github.com/liumency/SYSU-CD}{SYSU-CD}~\cite{shi2022sysucd}
  & 2021 & BCD & Opt. & Aerial & 0.5\,m & $256\!\times\!256$ & 20000 & Bi & Bldg., road, veg., water, marine construction \\
\href{https://github.com/S2Looking/Dataset}{S2Looking}~\cite{shen2021s2looking}
  & 2021 & BCD & Opt. & Satellite & 0.5--0.8\,m & $1024\!\times\!1024$ & 5000 & Bi & Buildings \\
\href{https://github.com/ShaoRuizhe/SUNet-change_detection}{HTCD}~\cite{shao2021sunet}
  & 2021 & BCD & Opt. & Sat.+UAV & VHR & Varies & 1293 & Bi & Buildings, roads, urban features \\
\href{https://captain-whu.github.io/SCD/}{SECOND}~\cite{yang2022second}
  & 2021 & SCD & Opt. & Aerial & 0.5--3\,m & $512\!\times\!512$ & 4662 & Bi & Ground, tree, low veg., water, bldg., playground \\
\href{https://github.com/Z-Zheng/ChangeOS}{ChangeOS}~\cite{zheng2021changeos}
  & 2021 & BDA & Opt. & Satellite & VHR & Varies & 2568 & Bi & No damage / minor / major / destroyed \\
\href{https://github.com/meiqihu/ACDA}{ACDA}~\cite{hu2021acda}
  & 2021 & BCD & HS & Air./Sat. & --- & Varies & 4 & Bi & Anomalous spectral changes \\
\href{https://registry.opendata.aws/spacenet/}{SpaceNet~7}~\cite{vanetten2021spacenet7}
  & 2021 & BCD & Opt. & Satellite & 4\,m & $1024\!\times\!1024$ & 101 AOIs & Multi & Buildings \\
\href{https://github.com/AmberHen/WHU-OPT-SAR-dataset}{WHU-OPT-SAR}~\cite{li2022mcanet}
  & 2022 & BCD & Cross & GF-2+S1 & 1--10\,m & Varies & 2695 & Bi & Farmland, urban, rural, water, forest, road \\
\href{https://github.com/liumency/CropLand-CD}{CLCD}~\cite{liu2022clcd}
  & 2022 & BCD & Opt. & GF-2 & 0.5--2\,m & $512\!\times\!512$ & 600 & Bi & Cropland \\
\href{https://github.com/Daisy-7/Hi-UCD-S}{Hi-UCD}~\cite{tian2022hiucd}
  & 2022 & SCD & Opt. & Aerial & 0.1\,m & $1024\!\times\!1024$ & 40800+ & Multi & Water, grass, forest, bare soil, bldg., greenhouse, road, bridge \\
\href{https://figshare.com/articles/figure/Landsat-SCD_dataset_zip/19946135/1}{Landsat-SCD}~\cite{yuan2022landsat}
  & 2022 & SCD & MS & Landsat & 30\,m & $256\!\times\!256$ & 8468 & Bi & Desert, farmland, bldg., water \\
\href{https://github.com/Lihy256/MSCDUnet}{MSBC}~\cite{li2022mscdunet}
  & 2022 & BCD & Multi & GF-2+S1+S2 & 0.8--10\,m & Varies & 4838 & Bi & Buildings \\
\href{https://mediatum.ub.tum.de/1650201}{DynamicEarthNet}~\cite{toker2022dynamicearthnet}
  & 2022 & SCD & MS & PlanetScope & 3\,m & $1024\!\times\!1024$ & 75 AOIs & Multi & Impervious, agriculture, forest, wetland, soil, water, snow \\
\href{https://github.com/fitzpchao/BANDON}{BANDON}~\cite{pang2023bandon}
  & 2023 & SCD & Opt. & Aerial & VHR & $2048\!\times\!2048$ & 2283 & Bi & Buildings \\
\href{https://github.com/VMarsocci/3DCD}{3DCD}~\cite{marsocci2023inferring3d}
  & 2023 & BCD & Opt. & Aerial & VHR & $512\!\times\!512$ & 472 & Bi & Buildings, roads, bridges \\
\href{https://github.com/CAU-HE/CMCDNet}{CAU-Flood}~\cite{he2023cauflood}
  & 2023 & BCD & Cross & Opt.+SAR & 10\,m & $256\!\times\!256$ & 2016 & Bi & Flood inundation \\
\href{https://github.com/AngieNikki/openWUSU}{WUSU}~\cite{shi2023wusu}
  & 2023 & SCD & Opt. & GF-2 & 0.8\,m & $512\!\times\!512$ & 10000+ & Multi & Bldg., road, cropland, forest, grass, river, lake, excavation, bare soil \\
\href{https://github.com/BinaLab/RescueNet-A-High-Resolution-Post-Disaster-UAV-Dataset-for-Semantic-Segmentation/tree/main}{RescueNet}~\cite{rahnemoonfar2023rescuenet}
  & 2023 & BDA & Opt. & UAV & VHR & $3000\!\times\!4000$ & 2396 & Single & Bldg., water, road, vehicle, tree, pool \\
\href{https://github.com/oshholail/EGY-BCD}{EGY-BCD}~\cite{holail2023afdenet}
  & 2023 & BCD & Opt. & Satellite & 0.25\,m & $256\!\times\!256$ & 6091 & Bi & Buildings \\
\href{https://github.com/lsmlyn/CropSCD}{CropSCD}~\cite{liu2024cropscd}
  & 2024 & SCD & Opt. & GF-2 & VHR & $256\!\times\!256$ & 4202 & Bi & Water, forest, plantation, grassland, impervious, greenhouse, road, bare soil \\
ClearSCD~\cite{tang2024clearscd}
  & 2024 & SCD & Opt. & Various & 0.5--3\,m & $512\!\times\!512$ & 4662 & Bi & Ground, tree, low veg., water, bldg., playground \\
\href{https://github.com/ya0-sun/PostEQ-SARopt-BuildingDamage}{QQB}~\cite{sun2024qqb}
  & 2024 & BDA & Cross & SAR+Opt. & VHR & $256\!\times\!256$ & 3200 & Bi & Buildings \\
\href{https://github.com/MarsZhaoYT/TSCD-Dataset}{TSCD}~\cite{zhao2024coud}
  & 2024 & BCD & Opt. & WorldView-2 & 0.5\,m & $256\!\times\!256$ & 5140 & Multi & Buildings, greenhouses \\
\href{https://github.com/circleLZY/MTKD-CD}{JL1-CD}~\cite{liu2025jl1cd}
  & 2025 & BCD & Opt. & JL-1 & 0.5--0.75\,m & $512\!\times\!512$ & 5000 & Bi & Bldg., road, PV, hardened, forest, grass, cropland, water \\
PRO-HRSCD~\cite{fang2025prohrscd}
  & 2025 & SCD & Opt. & Aerial (IGN) & 0.5\,m & $10000\!\times\!10000$ & 291 & Bi & Artificial, agriculture, forest, wetland, water \\
\href{https://github.com/StephenApX/MTL-TripleS?tab=readme-ov-file}{TripleS}~\cite{tan2025triples}
  & 2025 & SCD & Opt. & Various & 0.5--3\,m & $512\!\times\!512$ & 4662+ & Bi & Bare soil, water, bldg., farmland, veg., road \\
\href{https://github.com/ChenHongruixuan/BRIGHT}{BRIGHT}~\cite{chen2025bright}
  & 2025 & BDA & Cross & SAR+Opt. & 0.3--1\,m & $1024\!\times\!1024$ & 4538 & Bi & Buildings \\
CDFNet~\cite{wang2025cdfnet}
  & 2025 & BDA & Opt. & Satellite & VHR & Varies & 1500+ & Single & Multi-level building damage \\
\bottomrule
\end{tabular}
\end{adjustbox}
\vspace{2pt}

\parbox{\textwidth}{\scriptsize
\textbf{Abbreviations:} Opt.\ = Optical RGB; MS = Multispectral; HS = Hyperspectral; Cross = Cross-modal; Multi = Multi-modal fusion; GE = Google Earth; GF-2 = Gaofen-2; JL-1 = Jilin-1; S1/S2 = Sentinel-1/2; PV = Photovoltaic; Bi = Bi-temporal; Bldg.\ = Building(s); Veg.\ = Vegetation; VHR = Very High Resolution ($<$1\,m). Dataset names are hyperlinked to download pages where available.}
\end{table*}